\documentclass[journal]{IEEEtran}
\usepackage{amsmath,amsfonts}
\usepackage{algorithmic}
\usepackage{algorithm}
\usepackage{array}
\usepackage[caption=false,font=normalsize,labelfont=sf,textfont=sf]{subfig}
\usepackage{textcomp}
\usepackage{stfloats}
\usepackage{url}
\usepackage{verbatim}
\usepackage{graphicx}
\usepackage{cite}
\usepackage{balance}
\usepackage{tabularx}
\usepackage{booktabs}
\usepackage{multirow}
\usepackage{colortbl}
\begin{document}

\title{MEDN: Motion-Emotion Feature Decoupling Network for \\ Micro-Expression Recognition}

\author{Chenxing Hu, Kun Xie, Qiguang Miao*, \IEEEmembership{Senior Member, IEEE}, Ruyi Liu, Quan Wang, and Zongkai Yang

\thanks{This work was supported by The National Natural Science Foundations of China under grant No. 62272364, the provincial Key Research and Development Program of Shaanxi under grant No. 2024GH-ZDXM-47, the Higher Education Science Research Planning Project of China Association of Higher Education No. 24PG0101, the Fundamental Research Funds for the Central Universities, and the Postgraduate Innovation Fund of Xidian University No. YJSJ26014.}
\thanks{(* Corresponding author: Qiguang Miao.)}
\thanks{Chenxing Hu, Kun Xie, Qiguang Miao, and Ruyi Liu are with the School of Computer Science and Technology, Xidian University, Xi'an, Shaanxi 710071, China, Xi'an Key Laboratory of Big Data and Intelligent Vision, Xi'an, Shaanxi 710071, China and with Key Laboratory of Collaborative Intelligence Systems, Ministry of Education, Xidian University, Xi'an 710071, China (e-mail: huchenxing@stu.xidian.edu.cn; xiekun@xidian.edu.cn; qgmiao@xidian.edu.cn; ruyiliu@xidian.edu.cn).}
\thanks{Quan Wang is with the School of Computer Science and Technology, Xidian University, Xi'an, Shaanxi 710071, China (e-mail: qwang@xidian.edu.cn).}
\thanks{Zongkai Yang is with the National Engineering Research Center of Educational Big Data and the Faculty of Artificial Intelligence in Education, Central China Normal University, Wuhan 430079, China (e-mail: zkyang@mail.ccnu.edu.cn).}

}
\markboth{Journal of \LaTeX\ Class Files,~Vol.~14, No.~8, August~2021}%
{Shell \MakeLowercase{\textit{et al.}}: A Sample Article Using IEEEtran.cls for IEEE Journals}

\IEEEpubid{0000--0000/00\$00.00~\copyright~2021 IEEE}

\maketitle

\begin{abstract}
Unlike macro-expression, micro-expression does not follow a strictly consistent mapping rule between emotions and Action Units (AUs). As a result, some micro-expressions share identical AUs yet represent completely opposite emotional categories, making them highly visually similar. Existing micro-expression recognition (MER) methods mostly rely on explicit facial motion cues (e.g., optical flow, frame differences, AU features) while ignoring implicit emotion information. To tackle this issue, this paper presents a Motion-Emotion Feature Decoupling Network (MEDN) for MER. We design a dual-branch framework to separately extract motion and emotion features. In the motion branch, an AU-detection task restricts features to the explicit motion domain, and orthogonal loss is adopted to reduce motion–emotion feature coupling. For implicit emotion modeling, we propose a Sparse Emotion Vision Transformer (SEVit) that sparsifies spatial tokens to highlight local temporal variations with multi-scale sparsity rates. A Collaborative Fusion Module (CoFM) is further developed to fuse disentangled motion and emotion features adaptively. Extensive experiments on three benchmark datasets validate that MEDN effectively decouples motion and emotion features and achieves superior recognition performance, offering a new perspective for enhancing recognition accuracy and generalization.
\end{abstract}

\begin{IEEEkeywords}
Emotion Learning, Micro-expression, Feature Decoupling, Sparse Attention, Collaborative Fusion.
\end{IEEEkeywords}

\section{Introduction}
\IEEEPARstart{F}{acial} expressions contain abundant emotional information and play a crucial role in affective analysis\cite{ekman2003darwin}. Generally, facial expressions are categorized into macro-expressions and micro-expressions. In contrast to macro-expressions, micro-expressions are characterized by tiny motion amplitudes and extremely short durations, making them difficult to detect easily. Furthermore, macro-expressions can be deliberately faked to conceal genuine emotions, while micro-expressions are spontaneous emotional responses that usually occur involuntarily. Thus, micro-expression analysis has become an important means of revealing humans' true emotions\cite{porter2008reading} and is widely applied in fields such as medical care, criminal interrogation, and affective analysis\cite{o2009police},\cite{endres2009micro}. Manual micro-expression analysis poses a high threshold—even professionally trained experts can only identify less than 50\% of micro-expressions on average\cite{ben2021video}, which highlights the critical importance of developing automated micro-expression recognition algorithms. The task of micro-expression recognition aims to classify the time intervals where micro-expressions occur into specific emotional categories. However, the transient and subtle physical properties of micro-expressions lead to inherent challenges such as inconspicuous visual features and high susceptibility to noise interference, imposing great obstacles on the micro-expression recognition task.

\begin{figure}[!t]
  \centering
    \includegraphics[width=\linewidth]{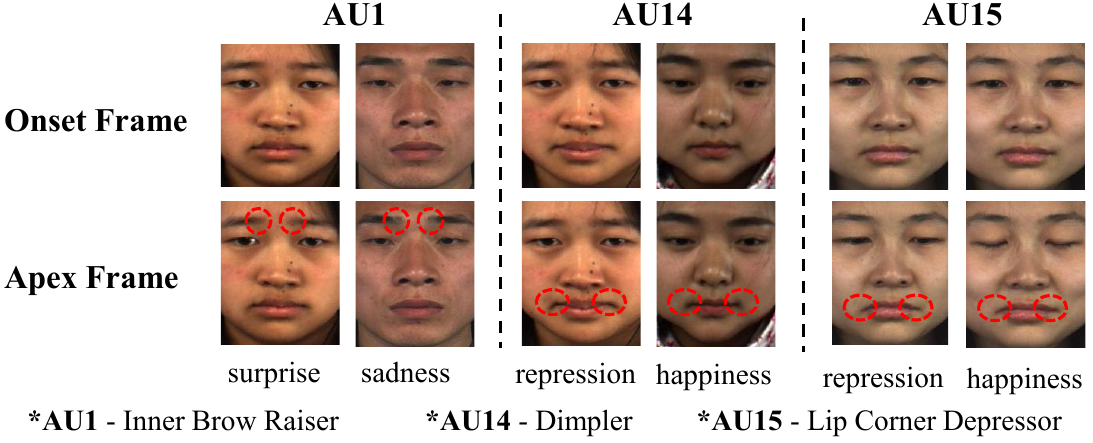}
    \caption{Some samples exhibit identical Action Unit activation patterns yet correspond to distinct emotion categories, characterized by highly similar visual motion features. This phenomenon can even occur in the same subject, highlighting the challenge of emotion discrimination relying solely on visual motion features.}
    \label{fig:micro_expression_examples}
\end{figure}

\IEEEpubidadjcol The Facial Action Coding System (FACS) has exerted a profound influence on facial expression analysis\cite{ekman1978facial}. Based on the anatomical structure of the human face, FACS encodes and classifies each visually distinguishable facial muscle movement, which is referred to as facial action units (AUs). Through psychological research, FACS provides an AU-emotion correspondence table for macro-expressions, enabling researchers to objectively label the emotional categories of macro-expressions according to AU combinations. For this reason, most datasets in micro-expression analysis also adopt AU labeling based on FACS. The key difference, however, is that the corresponding relationship between AU combinations and emotional categories in micro-expressions has not been strictly defined; researchers only take AUs as a reference rather than a decisive factor for emotional labeling.

\begin{figure*}[!t]  
    \centering       
    \includegraphics[width=1\textwidth]{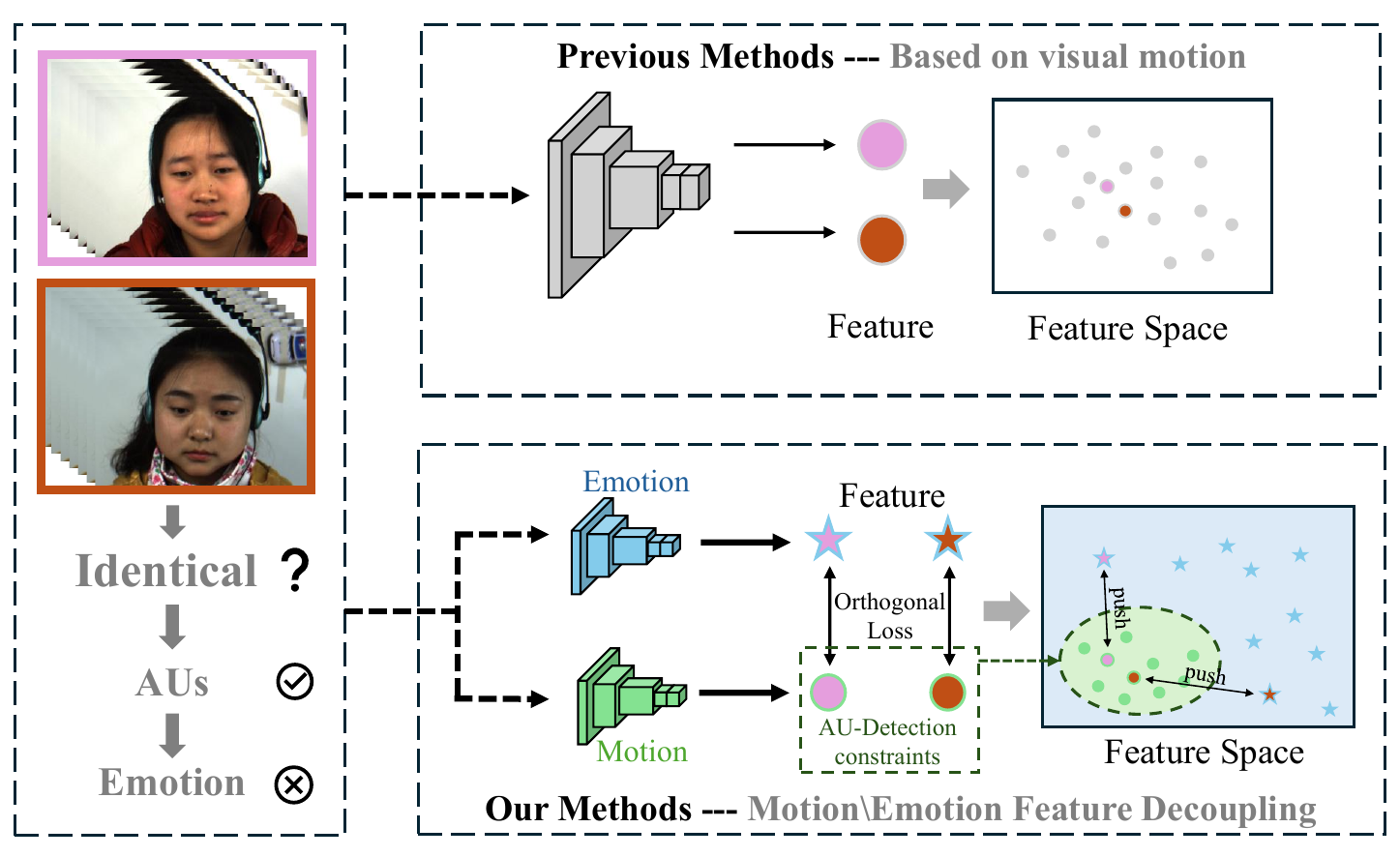}
    \caption{Previous MER methods focus on characterizing micro-expressions using explicit visual motion. In contrast, our method constrains motion feature extraction within a motion feature space via AU detection. Furthermore, an orthogonal loss is employed to prevent emotion features from attending to explicit motion features strongly correlated with AUs, thereby improving discrimination between micro-expressions that are visually similar but belong to different emotion categories.}
    \label{fig:MI}
\end{figure*}

AUs are highly consistent with the visual motion features of micro-expressions, providing explicit and reliable cues for visual motion feature extraction. For this reason, some micro-expression recognition methods\cite{xie2020auassisted,lei2021augcn,wang2024au} attempt to introduce AUs to assist in extracting micro-expression motion features, while others\cite{cai2024mfdan,liu2025merclip}, though not heavily reliant on AUs, only focus on extracting explicit visual motion features. These works have achieved considerable progress to a certain extent, yet they all overlook a crucial issue: the non-strict correspondence between AUs and emotional categories in micro-expressions means that AU motion features cannot play a decisive role in the final emotional classification of all samples. There exist samples in micro-expressions that share identical AUs but represent completely different emotional categories. This phenomenon is particularly significant in large-scale datasets and fine-grained recognition tasks. Such samples possess highly similar visual motion features, and over-reliance on AUs and explicit visual motion features will impair the generalization ability of models for the micro-expression recognition task. In this case, the implicit emotion features in these samples become the vital decisive factors. In summary, the features that determine the emotional categories of micro-expressions consist of two components: explicit visual motion features and implicit emotion features. Different samples vary in their feature dependence priorities. Thus, how to guide models to effectively extract and utilize these two types of features for micro-expression recognition has become a critical research problem.

To address the aforementioned challenges, this paper proposes the Motion-Emotion Feature Decoupling Network (MEDN). Specifically, two branches are constructed in the feature extraction stage. One branch takes the AU detection task as the supervision signal and focuses on extracting visual motion features highly correlated with AUs. The other branch concentrates on the motion evolution of micro-expression occurrence sequences. By enforcing feature orthogonality with the AU detection branch, this branch is constrained from focusing on motion features and is thus dedicated to extracting implicit emotion-related features. Furthermore, different samples rely on motion and emotion features with varying weights. The orthogonally decoupled features require more emphasis on exploring their collaborative information. To achieve better complementary fusion of the two-branch features, this paper designs a feature collaborative fusion module, which adaptively mines and fuses the effective collaborative information between motion features and emotion features.

In summary, the research contributions of this paper are summarized as follows:
\begin{itemize}
\item{To the best of our knowledge, this work is the first in the field of MER to address samples where AUs are consistent, but the corresponding emotion categories are inconsistent. Such samples exhibit high visual similarity, making accurate micro-expression recognition difficult using only visual features. The proposed method focuses on modeling both explicit motion features and implicit emotion features. By means of orthogonal disentanglement, the emotion feature extraction branch is constrained to disregard explicit motion features.}
\item{The Sparse Emotion Vision Transformer (SEVit) module is designed for emotion feature extraction, in which multi-scale sparse attention is utilized to break the model’s excessive dependence on static facial motion features.}
\item{To achieve better collaboration and complementarity between motion features and emotion features, a Collaborative Fusion Module (CoFM) is designed to adaptively mine the motion-emotion collaborative relationship in different samples, which significantly improves the recognition performance.}
\item{Extensive experiments are conducted on three micro-expression benchmark datasets, and the proposed method outperforms mainstream approaches on multiple tasks. Meanwhile, comprehensive ablation experiments also verify the effectiveness of the proposed method.}
\end{itemize}

The rest of this paper is organized as follows. Section~\ref{sec:2} reviews the existing research on MER and its limitations, and introduces related work on AU detection. Section~\ref{sec:3} describes our proposed MEDN method in detail. Section~\ref{sec:4} introduces the experimental settings and the adopted evaluation protocol. Section~\ref{sec:5} presents and analyzes the experimental results of the proposed method comprehensively. Section~\ref{sec:6} concludes the paper and discusses future work.

\section{Related Work}
\label{sec:2}

\subsection{Handcrafted Feature-based Methods}
Early works on MER primarily relied on handcrafted visual descriptive features to classify micro-expressions into emotional categories. Such methods exhibit strong interpretability and are less susceptible to the issue of small-scale micro-expression datasets.

Zhao et al.\cite{zhao2007dynamic} proposed the Local Binary Pattern from Three Orthogonal Planes (LBP-TOP), which extracts LBP features from three orthogonal planes to fuse spatial appearance and temporal motion information while drastically reducing computational complexity. For this reason, LBP-TOP has been widely adopted as a baseline method in numerous MER studies, and a variety of improved methods\cite{wang2015efficient},\cite{huang2016spontaneous} based on this algorithm have also been proposed subsequently.

Gradient-based features and optical flow features have also been widely applied. Representative include the histogram of gradients \cite{dalal2005histograms} and the histogram of image gradient orientation \cite{li2017towards}. Such features primarily focus on the edge variations of local facial textures, can accurately characterize the facial grayscale gradient differences induced by micro-expressions, and exhibit favorable geometric invariance.

However, handcrafted feature-based methods can only characterize shallow image semantic information and fail to capture motion semantics that are highly correlated with micro-expressions. Furthermore, manually designed feature extractors require sophisticated parameter tuning and detailed design, which results in extremely poor robustness and generalization ability of such methods.

\subsection{Deep Learning-based Methods}

Deep learning techniques have achieved remarkable progress in computer vision tasks. Deep neural networks enable the adaptive extraction of high-level facial features, and such methods have been widely applied in MER. Patel et al.\cite{patel2016selective} employed convolutional neural networks (CNNs) combined with transfer learning to transfer knowledge from macro-expressions to micro-expression recognition, representing a milestone work in deep learning-based micro-expression recognition. Subsequently, deep learning has gradually replaced handcrafted feature-based methods and become the mainstream approach.

\subsubsection{Static motion-based Methods}

Micro-expressions are characterized by being subtle and transient. Their occurrence sequences contain a large amount of redundant information, and such irrelevant information can severely negatively affect emotion recognition. Therefore, numerous studies adopt keyframe-based static motion information for micro-expression representation. Liong et al.\cite{gan2019off} computed the optical flow between the onset frame and apex frame as input to mitigate redundant information. This novel approach effectively integrates the superior motion representation capability of optical flow with the powerful deep feature learning capacity of CNNs. Quang et al.\cite{van2019capsulenet} proposed CapsuleNet, which performs recognition using only the apex frames of micro-expressions. Li et al.\cite{li2022mmnetmusclemotionguidednetwork} proposed a Muscle Motion-guided Network (MMNet), the main branch of the network extracts motion features via the frame difference between the onset frame and the apex frame, while the lightweight auxiliary branch takes the onset frame as input to generate facial position embeddings, which work with the main branch to map the motion features to specific facial regions.

Some researchers \cite{zhang2025dynamic,zhai2023feature,nguyen2023micron} have also introduced self-supervised techniques to enable models to learn facial motion features from inter-frame motion differences. Zhai et al.\cite{zhai2023feature} proposed FRL‑DGT, which pre-trains the network via a generative task‑driven displacement generation module. It generates adaptive dynamic displacement features from the onset and apex frames to avoid reliance on optical flow computation. In the classification stage, it combines full‑face feature extraction and AU‑based local feature extraction for micro‑expression classification. Similarly, Fan et al.\cite{Fan_2023_CVPR} proposed SelfME, a self-supervised learning framework. By learning motion patterns, the model can generate the motion field between two frames and perform micro-expression recognition and classification based on the obtained motion field.

Micro-expression recognition based on keyframes can significantly reduce information redundancy and eliminate interference from irrelevant factors. Therefore, the vast majority of existing works rely on keyframes by default \cite{khor2019dual,WANG2024128196,cai2024mfdan,10856363,Zhang_2025_ICCV}. However, this approach also has significant drawbacks. First, additional keyframe annotation is required in real-world scenarios, which severely limits its practicality. Second, keyframes only characterize the explicit motion differences of micro-expressions, ignoring the emotional cues implied in the temporal evolution process. The MEDN model proposed in this paper directly extracts features from the entire micro-expression sequence. This enables the model to capture deep emotional characteristics beyond static motion information, while also freeing it from reliance on keyframe annotation, thus yielding greater practical application value.

\subsubsection{AU-assisted Methods}

\begin{figure*}[!t]  
    \centering       
    \includegraphics[width=1\textwidth]{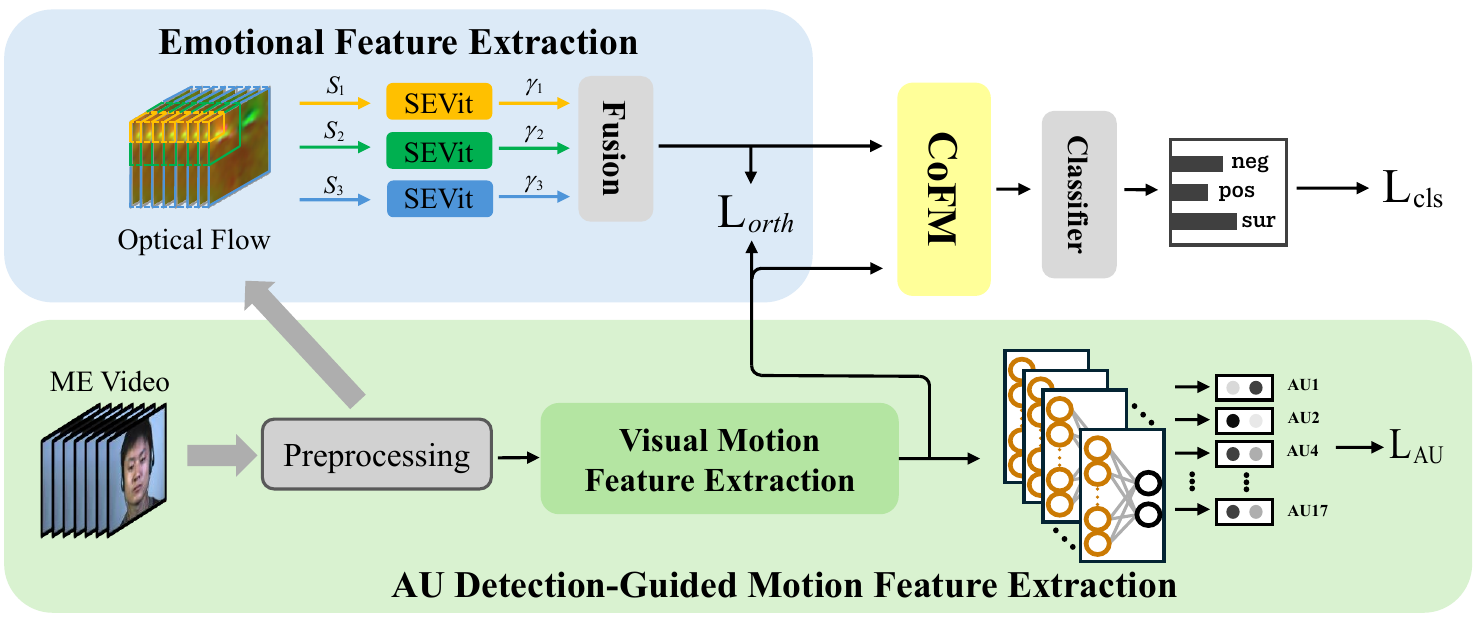}
    \caption{The architecture of our MEDN. For a micro-expression sequence, optical flow is first calculated frame by frame with respect to the first frame to obtain an optical flow sequence as input to the model. The optical flow sequence is then fed into the emotion feature extraction branch and the motion feature extraction branch for feature extraction. Among them, the motion feature extraction branch is supervised by the AU detection task as an auxiliary objective. The emotion feature extraction branch leverages our proposed SEViT to extract multi-scale features, and orthogonal decoupling with the motion feature extraction branch enables SEViT to focus more on implicit emotion features. Finally, the extracted motion and emotion features are fed into the CoFM module to achieve collaborative feature fusion, and the fused features are used for micro-expression recognition.}
    \label{fig:overall_architecture}
\end{figure*}

AU represents local facial movements and is highly correlated with the visual motion features of micro-expressions. Therefore, several studies employ AU information as guidance to assist MER.

Several works adopt Graph Neural Networks (GNN) to capture the interaction relationships among action units, and perform joint modeling based on the dynamic correlations between AUs and facial regions. Lo et al.\cite{9175230} proposed MER‑GCN, which leverages AU‑related annotations in the dataset to model the co‑occurrence relationships among action units. By employing Graph Convolutional Networks (GCN) to capture the dependencies between AUs, they integrate such relational information into MER, thereby boosting recognition performance. Lei et al.\cite{lei2021augcn} introduced an approach that unifies facial graph representation learning and facial Action Unit fusion. In this method, AU information is finely fused according to different facial regions, which enables effective capture of the subtle local features and node correlations of micro‑expressions, and facilitates a more thorough exploitation of AU cues.

Notably, some works directly introduce AU detection as a branch task into micro-expression recognition. Xie et al.\cite{xie2020auassisted} first employed AU detection as an auxiliary task for MER, making full use of facial muscle movement information to improve micro-expression recognition performance. Meanwhile, they generated realistic micro-expression sequences based on AU intensities, which effectively expanded the scale of micro-expression datasets. Zhou et al.\cite{10979653} obtained AU features supervised by the AU detection task, and fused the correlated features using GCN for MER. Wang et al.\cite{wang2024au} proposed a dual-branch network, where the two branches separately extract enhanced spatiotemporal features and dynamic AU features. Finally, the features from the two branches are fused via dot product to perform micro-expression classification.

However, existing AU-assisted methods only focus on using AU information to supplement or assist the extraction of visual motion features for micro-expressions, while ignoring affective cues beyond explicit motion features. Although our MEDN also takes AU detection as a separate branch, its purpose is to perform orthogonal disentanglement using motion strongly correlated features supervised by AU detection, so that the affective cue extraction branch no longer focuses on visual motion features.

\subsection{Facial AU Detection}

AU features are highly consistent with facial visual motion features and serve as an important representation for characterizing facial expression movements. Numerous studies have been conducted on AU detection\cite{li2017action,zhang2023detecting,chang2024facial,li2026hierarchical}. Early AU detection methods were mostly based on handcrafted features\cite{wang2013capturing,wu2015multi}, which generally lack generalization ability and scalability. Liu et al.\cite{liu2019relation} first proposed modeling AUs using GCN. This method extracts implicit representations of AU regions via an autoencoder, takes each implicit representation vector as a node input to the GCN, and determines the connection mode of GCN nodes according to the interaction relationships among AUs. Jacob et al.\cite{jacob2021facial} employed Transformers as the encoder to capture the interdependencies among various action units under different expressions. However, the above methods do not specifically focus on AU detection in micro-expressions.

Macro-expressions can obtain effective information directly from static facial appearance, whereas micro-expressions involve extremely subtle facial movements, requiring greater attention to inter-frame dynamic changes. Several studies\cite{li2021micro,zhang2021facial,li2021intra} have conducted AU detection specifically for micro-expressions, yet these methods still fail to fully exploit the rich information contained in facial dynamics. Varanka et al.\cite{varanka2023LED} proposed Learnable Eulerian Dynamics (LED), which combines automatic differentiation and linearized Eulerian video magnification to efficiently extract motion change representations from subtly varying frame sequences. Zhang et al.\cite{zhang2025regulatory} proposed the AC4AU framework for extracting dynamic features from frame sequences based on Regulatory Focus Theory, which integrates frequency-aware redundancy decomposition and an AU-specific expert routing mechanism.

\section{Our Method}
\label{sec:3}
In this section, we elaborate on the proposed MEDN method, whose overall architecture is illustrated in Figure \ref{fig:overall_architecture}. Firstly, we introduce how to extract explicit motion features. Next, we present the designed SEVit module and orthogonal decoupling method, which work together to accomplish emotion feature extraction. Subsequently, the proposed CoFM module is adopted to achieve collaborative feature fusion of motion and emotion features, and the fused features are used for MER. Finally, the overall loss function is described.

\subsection{Motion Feature Extraction}

\begin{figure}
    \centering
        \includegraphics[width=\linewidth]{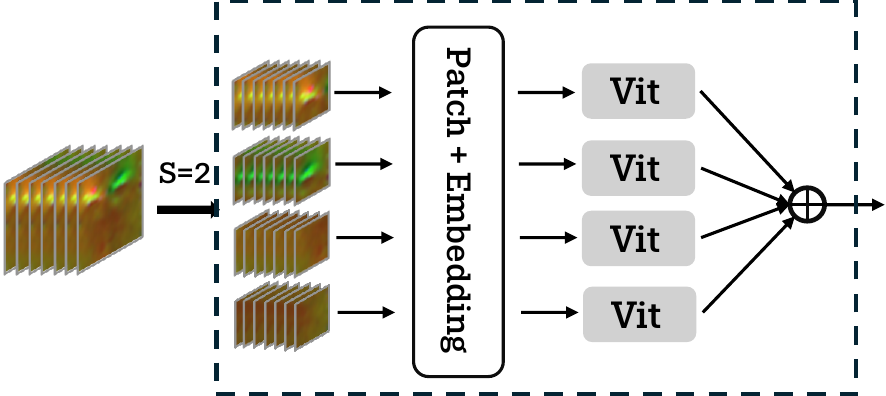}
        \caption{The details of the proposed SEVit module. When the sparsity rate $S=2$, the optical flow sequence is divided into 4 parts along the spatial dimension while the temporal dimension remains continuous. Then, patch partitioning, encoding, and attention computation are performed separately in the manner of ViT. Finally, the features are concatenated.}
        \label{fig:SEVit}
\end{figure}

The motion feature extraction branch is designed to extract explicit visual motion features. Given the high correspondence between such features and the explicit motions of AU, which provide reliable and interpretable guidance for feature extraction, this branch adopts the AU detection task as an auxiliary supervision signal to guide the model in learning motion-related features.

In this work, the AU detection branch is implemented based on the Learnable Eulerian Dynamics (LED) model proposed by Varanka et al.~\cite{varanka2023LED}, which has demonstrated strong performance in AU-related feature extraction. The LED model leverages both local and global facial features to enhance AU detection, making it well-suited for capturing the explicit motion patterns associated with Action Units.

The AU detection task is formulated as a multi-label classification problem, where each AU is treated as an independent binary classification sub-task. To supervise this task, we adopt the standard multi-task cross-entropy loss, which is widely used for multi-label AU detection. The loss is computed as the sum of cross-entropy losses over all target AUs:

\begin{equation}
\mathcal{L}_{\mathrm{au}} = \sum_{i=1}^{N_{\mathrm{AU}}} \mathcal{L}_{\mathrm{CE}}(p_i, y_i),
\end{equation}
where $N_{\mathrm{AU}}$ is the number of Action Units, $\mathcal{L}_{\mathrm{CE}}(\cdot,\cdot)$ denotes the cross-entropy loss, $p_i$ is the predicted probability for the $i$-th AU, and $y_i$ is its ground-truth label (0 or 1).

In practice, the cross-entropy loss for each AU sub-task quantifies the discrepancy between the model’s predicted likelihood of the AU being activated/non-activated and the true label, ensuring that the model is penalized for misclassifying the activation state of each AU. By summing these losses across all AUs, the multi-task loss enforces the model to simultaneously learn discriminative features for all target AUs, rather than optimizing for a single AU in isolation.

\subsection{Emotion Feature Extraction}

\subsubsection{Sparse Emotion Vision Transformer}

Traditional global attention mechanisms \cite{dosovitskiy2020vit} often lead models to overly focus on explicit static visual semantics. For the image manipulation localization task (IML), Su et al. \cite{su2025can} decomposed the global attention mechanism into sparse attention, enabling the model to better capture region-independent features within individual frames. The emotional cues of micro-expressions are often implicit in the temporal motion variations, rather than in static motion deformations themselves. Therefore, the model should focus more on local feature changes along the temporal dimension. Inspired by \cite{su2025can}, we design the Sparse Emotion Vision Transformer (SEVit) as the basic module for emotion feature extraction. SEVit can break the dominance of static visual information in micro-expression sequence feature extraction and force the model to mine fine-grained emotional cues within the local temporal feature flow.

As shown in Figure \ref{fig:SEVit}, given a micro-expression sequence tensor $\mathbf{X} \in \mathbb{R}^{T \times H \times W \times C}$ (where $T$ denotes the number of time steps, $H/W$ represent the height/width of the feature map, and $C$ is the number of channels), we no longer simply perform patch partitioning and global attention calculation in a naive manner. Instead, we first divide the features sequentially into $s^2$ non-overlapping sub-tensors $\{\mathbf{N}_i\}_{i=1}^{s^2}$ along the spatial dimension according to a sparsity rate $s$ (degenerating to the standard global attention when $s=1$), where each $\mathbf{N}_i \in \mathbb{R}^{T \times H/s \times W/s \times C}$. Then, self-attention is computed independently within each $\mathbf{N}_i$. This approach can forcibly suppress the model's over-reliance on static spatial features and drive it to focus more on temporal variation information within local regions.

\begin{equation}
\text{SEVit}(X, s) = \bigoplus_{i=1}^{s^2} \text{SelfAttention}(N_i),
\end{equation}
where $\bigoplus$ denotes the operation of concatenating the outputs from all self-attention modules.

Furthermore, we employ multi-scale feature fusion with parallel sparse rates. This prevents the destruction of the internal spatial correlations within critical local regions when dividing tensors using a single sparse rate, thereby improving the generalization ability of the model. Specifically, given a set of sparsity rates (where $K$ is the number of sparsity rates), the feature extracted under the $k$-th sparsity rate is denoted as:

\begin{equation}
F_k = \text{SEVit}(X, s_k), k=1,\ldots, K.
\end{equation}

Then, a learnable fusion weight $\gamma_k$ is introduced for each feature $F_k$ to balance the multi-scale emotion features under different sparsity rates. The final fused feature is given by:
\begin{equation}
F_{\text{e}} = \sum_{k=1}^{K} \gamma_k F_k.
\end{equation}

\subsubsection{Orthogonal Loss}

To encourage the model to focus more on affective features rather than explicit motion features, we employ orthogonal loss to disentangle affective features from motion features supervised by AU detection in the feature space. This disentanglement guides the model to capture deep affective cues beyond explicit motion. By constructing a dedicated constraint term, orthogonal loss effectively decouples implicit affective features and explicit motion features, reduces information redundancy between the two types of features, and ensures that the model can fully mine the core semantic cues related to affective expression.

Specifically, the extracted motion and affective features are denoted as $F_m$ and $F_e$ ($F_m, F_e \in \mathbb{R}^{1\times D} $), where $D$ denotes the feature dimension. To eliminate the scale discrepancy between the two types of features and ensure the rationality of subsequent calculations, we first perform L2 normalization on the features. The specific formulation is given as follows.

\begin{equation}
    \hat{F}_m = \frac{F_m}{\|F_m\|_2 + \epsilon}, \quad
    \hat{F}_e = \frac{F_e}{\|F_e\|_2 + \epsilon},
\end{equation}
where $\hat{F}_m,\hat{F}_e$ denote the normalized feature vectors and $\epsilon$ is a small constant for numerical stability.

To quantify the degree of linear correlation between features, we calculate the projection coefficient between features, which is obtained by taking the dot product of the two types of normalized features. The specific expression is given as follows:

\begin{equation}
    \alpha = \langle  \hat{F}_m, \hat{F}_e \rangle.
\end{equation}

Based on the above projection coefficient $\alpha$, the orthogonal components of $\hat{F}_m$ and $\hat{F}_e$ with respect to each other can be solved respectively, thereby stripping the linear correlation between the two types of features and preserving their respective unique information. The specific calculations are given as follows:

\begin{equation}
    F_m^\perp = \hat{F}_m - \alpha \cdot \hat{F}_e, \quad
    F_e^\perp = \hat{F}_e - \alpha \cdot \hat{F}_m.
\end{equation}

The core idea of orthogonal loss is to achieve feature disentanglement by constraining the orthogonal components of the two types of features to be as independent of each other as possible. It is defined as the mean square value of the inner product of the two types of features after orthogonalization, and scaled by the feature dimension to balance the relationship between the loss scale and the feature dimension. Therefore, the final specific loss function is expressed as follows:

\begin{equation}
    \mathcal{L}_{\text{orth}} = D \cdot \langle  F_m^\perp, F_e^\perp \rangle^2.
\end{equation}

This loss function effectively strips redundant information from the two types of features by minimizing the sum of squared inner products between orthogonal components, ensuring that emotion features can break away from the dependence on explicit motion features, preserve the unique semantic information inherent to emotional expressions, and further guide the model to focus on deep emotional cues beyond explicit motions.

\subsection{Collaborative Fusion Module}

\begin{figure}
    \centering
        \includegraphics[width=\linewidth]{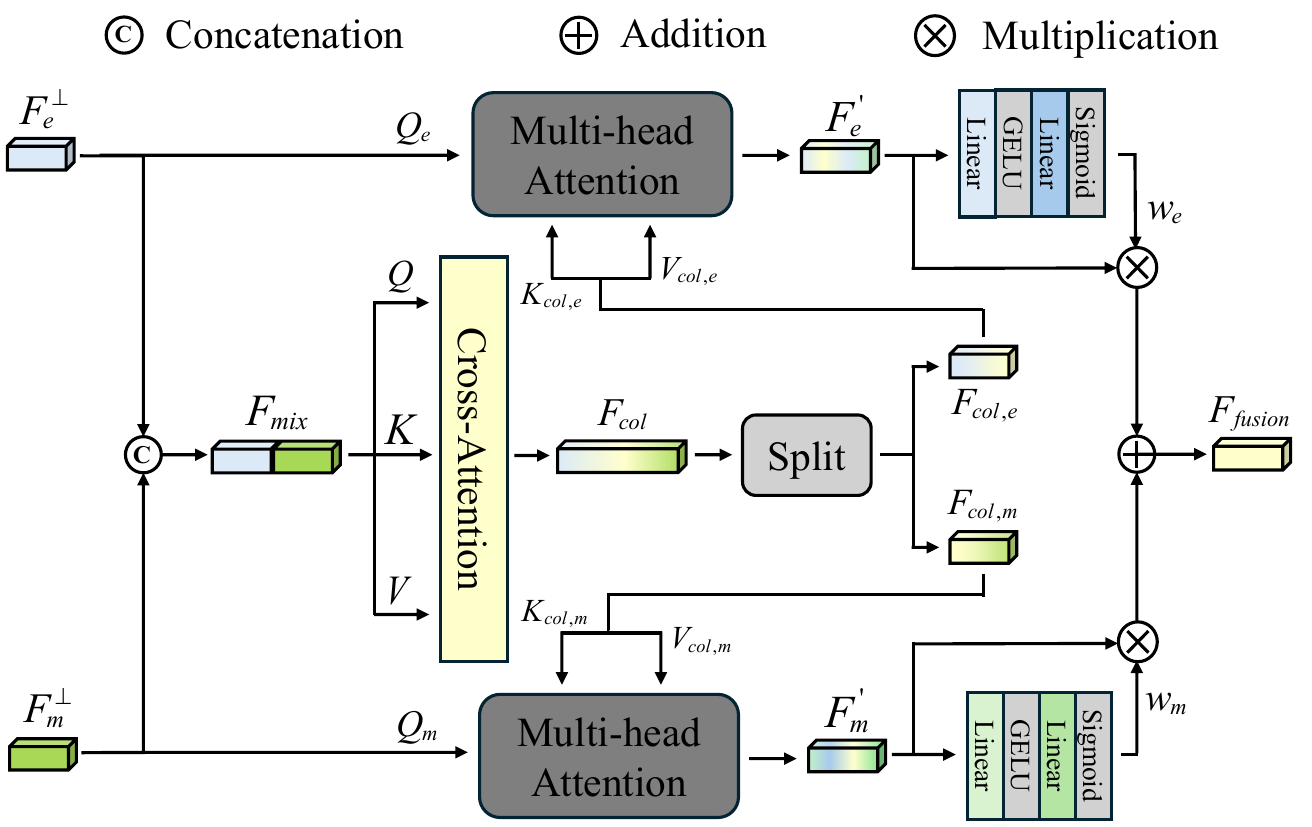}
        \caption{The details of the proposed CoFM module. The CoFM module aims to model the collaboration between the orthogonalized features. Then, multi-head attention is computed using the queries and keys generated from the original features and the features after collaborative modeling. Finally, the collaboration-enhanced aggregated features are obtained via adaptive weight fusion.}
        \label{fig:CoFM}
\end{figure}

Different samples exhibit distinct dependencies on motion and emotion features; thus, we require a fusion mechanism to adaptively balance their contributions. Furthermore, since motion and emotion features are constrained to separate subspaces by the orthogonal loss, they present substantial feature-level disparities. Direct fusion of the two features may lead to semantic mismatches. To address this issue, we propose the Collaborative Fusion Module (CoFM), which aims to explore the collaborative patterns between features from the two independent branches and learn an optimal complementary fusion strategy for collaborative features, as detailed in Figure \ref{fig:CoFM}. Unlike existing correlation-based methods, CoFM focuses on modeling collaborative interactions between the orthogonalized motion and emotion features, guiding them to complement each other and improve micro-expression recognition performance. Specifically, we first model the collaborative information of the two features by concatenating the orthogonally decoupled motion features $F_m^\perp$ and emotion features $F_e^\perp$ along the feature dimension to construct the mixed feature tensor $F_{mix}$, expressed as:

\begin{equation}
    F_{mix} = \text{Concat}(F_m^\perp, F_e^\perp).
\end{equation}

To project the mixed feature tensor into a feature space suitable for collaborative modeling, linear layers are employed to map it to the query vector $Q$, key vector $K$, and value vector $V$, respectively:

\begin{equation}
    Q,K,V = \text{Linear}(F_{mix}).
\end{equation}

To accurately capture the fine-grained collaborative relationships between the two branches, we explore the collaborative patterns within the feature subspace via the attention mechanism, with the specific implementation detailed as follows:

\begin{equation}
    F_{col} = \text{softmax}\left( \frac{QK^\top}{\sqrt{d_k}} \right) V,
\end{equation}
where $\sqrt{d_k}$ is the scaling factor, denotes the mixed vector after collaborative modeling, and $F_{col}$ is finally split into two mixed sub-vectors $F_{col,m}$ and $F_{col,e}$. This operation enables modeling of collaboratively enhanced features after orthogonal decoupling in separate feature spaces, allowing the model to adaptively align information according to the feature collaboration relationships.

The collaboratively modeled features $F_{col,m}$ and $F_{col,e}$ contain collaborative information between motion and emotion. To generate the final collaboratively enhanced features, we map the original orthogonalized features $F_m^\perp$ and $F_e^\perp$ to query vectors $Q_m$,$Q_e$. Meanwhile, $F_{col,m}$ and $F_{col,e}$  are mapped to key vectors $K_{col,m}$,$K_{col,e}$ and value vectors $V_{col,m}$,$V_{col,e}$, based on which multi-head attention is computed for the motion and emotional branches separately to obtain the collaboratively enhanced features $F'_m$ and $F'_e$. Lightweight sequential networks are adopted to generate fusion weights $w_m$ and $w_e$:

\begin{equation}
    w_m = \text{Sigmoid}(W_1(\text{GELU}(W_2(F'_m)))),
\end{equation}
\begin{equation}
    w_e = \text{Sigmoid}(W_3(\text{GELU}(W_4(F'_e)))),
\end{equation}
where $W_1$,$W_2$,$W_3$,$W_4$ are learnable weight matrices. To avoid weight imbalance, the weights $w_m$and$w_e$ are further normalized. The final feature fusion process is as follows:

\begin{equation}
    F_{fusion}=w_mF'_m + w_eF'_e.
\end{equation}

The fused feature $F_{fusion}$ for final classification prediction. The CoFM module adapts to feature fusion after orthogonal decoupling, dynamically adjusts the contributions of the two branches according to the specific features of the input sample, and fully explores and integrates the collaborative and complementary information between motion and emotion features.

\subsection{Full Loss}

To alleviate the severe sample imbalance problem of micro-expression data, we adopt Focal Loss\cite{cai2024mfdan,lin2017focal} as the classification loss, and its formula is as follows:
\begin{equation}
\mathcal{L}_{cls}(p_t) = -(1-p_t)^\gamma \log(p_t).
\end{equation}

In this formula, \( p_t \) denotes the predicted probability of each category, and \( \gamma \) is a constant. Focal Loss adopts a dynamic weighting approach, which enables models to pay more attention to difficult-to-categorize samples. This characteristic effectively avoids the problem of models overfitting to easily classified samples.

In the proposed MEDN model, the total loss function is composed of three terms: the emotion classification loss $\mathcal{L}_{\mathrm{cls}}$, the AU detection loss $\mathcal{L}_{\mathrm{au}}$, and the orthogonal disentanglement loss $\mathcal{L}_{\mathrm{orth}}$. The overall loss is formulated as:
\begin{equation}
\mathcal{L}_{total} = \mathcal{L}_{\mathrm{cls}} + \lambda_{au} \mathcal{L}_{\mathrm{au}} + \lambda_{orth} \mathcal{L}_{\mathrm{orth}},
\end{equation}
where $\lambda_{au}$ and $\lambda_{orth}$ are hyperparameters that control the relative contributions of the AU detection and orthogonal losses.

\section{EXPERIMENTAL SETUP}
\label{sec:4}
In this section, we elaborate on the configuration and preparation of the experiments, including the dataset, preprocessing procedures, and the evaluation protocols adopted in the experiments.

\subsection{Datasets}

We validate our method on three mainstream datasets commonly used for micro-expression recognition. All these datasets contain authentic emotion labels and AU labels that our experiments rely on.

\textbf{CASMEII}\cite{yan2014casme2} is a widely used spontaneous micro-expression dataset for micro-expression recognition research. The dataset contains 255 micro-expression videos from 26 subjects, selected from nearly 3,000 elicited facial movements. Each video has a resolution of 280×340 and a frame rate of 200 FPS. In our experiments, we mainly use 145 samples for the three-class classification task.

\textbf{SAMM}\cite{davison2016samm} comprises 159 ME videos collected from 29 subjects (covering 13 ethnicities) under controlled lighting conditions without flickering. Recorded at 200 FPS with a facial resolution of 400$\times$400. In our experiments, we mainly use 133 samples for the three-class classification task.

\textbf{CAS(ME)$^3$}\cite{li2022casme33} mainly consists of three parts, among which Part A is the primary subset adopted in our experiments. This part comprises 860 valid micro-expression samples with complete annotations, and the videos were captured at a frame rate of 30 fps. The samples are annotated with seven emotion labels, including happiness, disgust, surprise, fear, anger, sadness, and others. For this dataset, we conduct comparative experiments on 3-class, 4-class, and 7-class classification tasks.

For the CASMEII and SAMM datasets, we adopt the Composite Database Evaluation (CDE) protocol proposed in MEGC2019 to unify the emotion categories into three classes: \textbf{Negative} \{"Repression", "Anger", "Contempt", "Disgust", "Fear", "Sadness"\}, \textbf{Positive} \{"Happiness"\}, and \textbf{Surprise} \{"Surprise"\}. Details are shown in Table \ref{tab:dataset}. For the CAS(ME)$^3$ dataset, we also divide the labels into three categories according to the above criteria. For the four-category task, we add an \textbf{Other} class, while for the seven-category task, we adopt the original emotion labels.

Due to the larger scale of the CAS(ME)$^3$ dataset, the issue of samples sharing the same AUs but belonging to different emotion categories becomes much more prominent. Table \ref{tab:dataset_casme33} presents the total number of samples, the number of hard samples, and the proportion of hard samples under different classification tasks on CAS(ME)$^3$. It can be seen from the table that more than 80\% of the samples have other easily confused samples with consistent AUs but inconsistent emotion labels. Accordingly, our ablation studies are mainly conducted on the CAS(ME)$^3$ dataset to verify the effectiveness of our proposed method.

\subsection{Data Pre-processing}

\begin{table}[]
\centering
\setlength{\tabcolsep}{16pt}
\caption{The number of videos in each category for the CASMEII and SAMM used in the 3-class classification task.}
\renewcommand{\arraystretch}{1.2}
\begin{tabular}{cccc}
\hline
Label                               & CASMEII       & SAMM        & Combined  \\ \hline
\multicolumn{1}{c|}{Positive}       & 32            & 26          & 58        \\
\multicolumn{1}{c|}{Negative}       & 88            & 92          & 180       \\
\multicolumn{1}{c|}{Surprise}       & 25            & 15          & 40        \\ \hline
\multicolumn{1}{c|}{Total}          & 145           & 133         & 278       \\ \hline
\end{tabular}
\label{tab:dataset}
\end{table}

\begin{table}[]
\centering
\setlength{\tabcolsep}{14pt}
\caption{The number of videos under different categories in the CAS(ME)$^3$ dataset. "Hard" denotes the number of hard samples (samples with the same AUs but different emotional categories), and "Proportion" represents the proportion of hard samples.}
\renewcommand{\arraystretch}{1.2}
\begin{tabular}{cccc}
\hline
Task                               & Total       & Hard       & Proportion     \\ \hline
\multicolumn{1}{c|}{3-class}       & 699         & 594        & 84.98\%        \\
\multicolumn{1}{c|}{4-class}       & 860         & 755        & 87.79\%        \\
\multicolumn{1}{c|}{7-class}       & 860         & 784        & 91.16\%        \\ \hline
\end{tabular}
\label{tab:dataset_casme33}
\end{table}

Our MEDN model takes video sequences as input. Thus, we uniformly sample $T=8$ frame images from the micro-expression occurrence interval of each sample to form micro-expression sequences. We detect the 68 facial landmarks and align for each frame image using the Dlib\cite{king2009dlib} toolkit, and then crop all image frames to a size of 3$\times$144$\times$144. Next, the TV-L1\cite{zach2007tvl1} optical flow method is adopted to calculate the optical flow between each frame and the first frame in the sequence; the optical flow in the horizontal and vertical directions is treated as two separate channels, yielding an optical flow sequence with a frame length of $T-1$ and a size of 2$\times$144$\times$144. Ultimately, the size of a single sample input to the MEDN model is $(T-1, C, H, W)$, where $T-1=7$, $C=2$, and $H=W=144$.

\subsection{Evaluation Protocols and Metrics}
Following existing research in the field of micro-expression recognition, we adopt the Leave-One-Subject-Out (LOSO) strategy for model evaluation. This strategy takes samples from a single individual subject as the test set for each round of cross-validation, and the final evaluation metrics are calculated by aggregating all prediction results from all rounds.

Since class distribution imbalance is a prevalent issue in micro-expression datasets, using the traditional Accuracy metric alone is insufficient to comprehensively evaluate model performance. For this reason, we select the Unweighted Average Recall (UAR) and Unweighted F1-score (UF1) as the core evaluation metrics to ensure the comprehensiveness of the experimental results.
We can calculate as follows:

\begin{equation}
\text{UF1} = \frac{1}{C}\sum_{i=1}^{C}\frac{2\times\text{TP}_i}{2\times\text{TP}_i + \text{FP}_i + \text{FN}_i},
\end{equation}
\begin{equation}
\text{UAR} = \frac{1}{C}\sum_{i=1}^{C}\frac{\text{TP}_i}{N_i},
\end{equation}
where $C$ is the number of categories, $\mathrm{TP}_i$, $\mathrm{FP}_i$, and $\mathrm{FN}_i$ denote the number of true positive, false positive, and false negative samples for category $i$, respectively, and $N_i$ is the total number of samples in category $i$.

\subsection{Implementation Details}

\begin{table*}[!t]  
    \centering  
    \caption{Comparison with the state-of-the-art methods on CASMEII and SAMM for the three-category classification task. The best results are highlighted in bold, while the second-best results are marked with an underline. "Apex Frame" indicates whether the method relies on apex frame annotations, and "Average" denotes the average results over the two datasets.}  
    \renewcommand{\arraystretch}{1.2}
    \begin{tabular*}{\textwidth}{@{\extracolsep{\fill}}cc|cccccc@{}}
        \toprule  
        \multirow{2}{*}{Methods} & \multirow{2}{*}{Apex Frame} & \multicolumn{2}{c}{CASMEII} & \multicolumn{2}{c}{SAMM} & \multicolumn{2}{c}{Average}  \\
        & & UF1 & UAR & UF1 & UAR & UF1 & UAR \\
        \midrule  
        CapsuleNet(2019)\cite{van2019capsulenet}   & $\surd$  & 0.7068 & 0.7018 & 0.6209 & 0.5989 & 0.6639 & 0.6504  \\
        STSTNet(2019)\cite{8756567}                & $\surd$  & 0.8382 & 0.8686 & 0.6588 & 0.6810 & 0.7489 & 0.7748  \\
        Neural(2019)\cite{liu2019neural}           & $\surd$  & 0.8293 & 0.8209 & 0.7754 & 0.7152 & 0.8024 & 0.7681  \\
        MTMNet(2020)\cite{xia2020learning}         & $\surd$  & 0.8700 & 0.8720 & 0.8250 & 0.8190 & 0.8475 & 0.8455  \\
        AUGCN(2021)\cite{lei2021augcn}             & $\surd$  & 0.8798 & 0.8710 & 0.7751 & 0.7890 & 0.8271 & 0.8300  \\
        MERASTC(2021)\cite{gupta2021merastc}       & $\times$ & 0.9330 & \textbf{0.9500} & 0.8300 & \underline{0.8460} & 0.8815 & \textbf{0.8980}  \\
        ME-PLAN(2022)\cite{zhao2022me}             & $\times$ & 0.8941 & 0.8962 & 0.7358 & 0.7687 & 0.8149 & 0.8325  \\
        GLEFFN(2023)\cite{guo2023gleffn}           & $\times$ & 0.8825 & 0.9110 & 0.7458 & 0.7843 & 0.8142 & 0.8477  \\
        Micron-BERT(2023)\cite{nguyen2023micron}   & $\surd$  & 0.9034 & 0.8914 & -      & -      & -      & -       \\
        MFDAN(2024)\cite{cai2024mfdan}             & $\surd$  & 0.9134 & 0.9326 & 0.7871 & 0.8196 & 0.8503 & 0.8761  \\
        MPFNet-P(2025)\cite{ma2025multi}           & $\surd$  & 0.8790 & 0.8950 & 0.7910 & 0.8260 & 0.8350 & 0.8605  \\
        MER-CLIP(2025)\cite{liu2025merclip}        & $\times$ & \textbf{0.9409} & \underline{0.9487} & 0.8321 & 0.8434 & 0.8865 & \underline{0.8961}  \\
        SODA4MER(2025)\cite{zhang2025dynamic}      & $\times$ & 0.8870 & 0.8809 & -      & -      & -      & -       \\
        GAMDSS(2026)\cite{11424978}                & $\surd$  & \underline{0.9362} & -      & \underline{0.8523} & -      & \underline{0.8943} & -       \\ 
        MPFNet-C(2026)\cite{11192760}              & $\surd$  & 0.9110 & 0.9230 & 0.7950 & 0.8390 & 0.8530 & 0.8810  \\ \hline
        MEDN(ours)                                 & $\times$ & 0.9118 & 0.8912 & \textbf{0.8835} & \textbf{0.8679} & \textbf{0.8977} & 0.8796  \\
        \bottomrule  
    \end{tabular*}
    \footnotesize  
    \label{tab:SOTA_SAMM_CASMEII}  
\end{table*}

\begin{table}[]
\centering
\caption{Comparison of experimental results with other methods for CAS(ME)$^3$ PartA}
\renewcommand{\arraystretch}{1.2}
\begin{tabularx}{\linewidth}{@{\extracolsep{\fill}}cccc}
\hline
Methods                              & Classes & UF1             & UAR             \\ \hline
STSTNet(2019)\cite{8756567}                 & 3       & 0.3795          & 0.3792          \\
Micron-BERT(2023)\cite{nguyen2023micron}    & 3       & 0.5604          & 0.6125          \\
HTNet(2024)\cite{WANG2024128196}            & 3       & 0.5767          & 0.5415          \\
Lite-Point-GCN(2025)\cite{10856363}         & 3       & \underline{0.6819}          & \textbf{0.7412} \\
MEDN(ours)                                  & 3       & \textbf{0.7568} & \underline{0.7294}          \\ \hline
Micron-BERT(2023)\cite{nguyen2023micron}    & 4       & 0.4718          & 0.4913          \\    
Lite-Point-GCN(2025)\cite{10856363}         & 4       & \underline{0.4764}          & \underline{0.5366}    \\
MEDN(ours)                                  & 4       & \textbf{0.6514} & \textbf{0.6102}             \\ \hline
Micron-BERT(2023)\cite{nguyen2023micron}    & 7       & 0.3264          & 0.3254          \\    
Lite-Point-GCN(2025)\cite{10856363}         & 7  & \underline{0.3564} & \underline{0.4159}    \\
MEDN(ours)                                  & 7       & \textbf{0.4906} & \textbf{0.4631} \\ \hline
\end{tabularx}
\label{tab:SOTA_CASME33}
\end{table}

All experiments in this study were conducted on the NVIDIA RTX 5880 Ada Generation graphics card, and the model was implemented based on the PyTorch\cite{paszke2019pytorch} framework. The $\gamma$  of Focal Loss was set to 2.0, and the batch size was configured as 128. The initial learning rate of the entire model was set to 0.0001. To stabilize the extraction of explicit motion features, a learning rate decay strategy was applied to the motion feature extraction branch, with the learning rate decayed to 0.1 times the original value every 100 epochs. The AdamW optimizer was adopted for model training, which fixes the defect of unstable weight decay effect of Adam in adaptive optimization by decoupling weight decay, with the weight decay set to 0.001.

\section{Results and Discussion}
\label{sec:5}
In this section, we compare the proposed MEDN model with state-of-the-art methods on multiple ME datasets, and MEDN achieves strong competitive performance on various tasks. We also conduct sufficient ablation experiments to verify the effectiveness of each module in the model.

\subsection{Comparison with State-of-the-Art Methods}

Table \ref{tab:SOTA_SAMM_CASMEII} presents the quantitative comparison between the proposed MEDN and state-of-the-art methods on the CASMEII and SAMM datasets. It can be observed that MEDN achieves competitive performance on both datasets. In particular, our method obtains significant advantages on the more challenging SAMM dataset: it outperforms the second-best method by 3.7\% in UF1 and 2.6\% in UAR, achieving comparable results to those on the CASME II dataset. This verifies the robustness and generalization ability of the proposed method in challenging scenarios. Among the methods that do not rely on the apex frame, our method also achieves the best performance on the CASME II dataset.

We conducted experiments on the CAS(ME)$^3$ dataset for 3-class, 4-class, and 7-class classification tasks, with the results presented in Table \ref{tab:SOTA_CASME33}. Compared with current mainstream methods, MEDN achieves significant performance improvements, especially in the finer-grained 4-class and 7-class tasks. MEDN improves the UF1 and UAR metrics by approximately 0.15, while suffering less performance degradation in even finer-grained emotion classification tasks. This demonstrates that in scenarios involving finer-grained emotion classification with highly consistent AUs but distinct emotions, the disentanglement of movement and emotion features by MEDN can significantly enhance the recognition accuracy and robustness of the model.

Unlike macro-expressions, micro-expressions contain a large number of visually confusing samples with consistent AU annotations within the same emotion category. Such samples pose challenges to the feature extraction paradigm in micro-expression recognition. Excessive focus on explicit motion features often makes the model more prone to confusing these ambiguous samples. Compared with other models, the proposed method guides the model to pay more attention to the implicit emotion features hidden under explicit motion features through orthogonal decoupling of motion and emotion features. It can adaptively exploit the collaborative relationship between motion and emotion features, achieving significant improvements in recognition accuracy.

\subsection{Ablation Study}

\begin{table*}[!t]  
    \centering  
    \caption{Ablation study of key components in the proposed MEDN method. $\surd$ indicates that the component is used, and $\times$ indicates that the component is not used.}  
    \renewcommand{\arraystretch}{1.2}
    \begin{tabular*}{\textwidth}{@{\extracolsep{\fill}}c|cccc|cc@{}}
        \toprule  
        \multirow{2}{*}{Number} & \multirow{2}{*}{SEVit} & \multirow{2}{*}{Multi-scale Sparse Rates}  & \multirow{2}{*}{Orthogonal Loss} & \multirow{2}{*}{CoFM} & \multicolumn{2}{c}{CAS(ME)$^3$-3class($\downarrow$)} \\ 
        & & & & & UF1 & UAR \\ \hline
        No.1 & $\times$      & $\times$    & $\surd$    & $\surd$      & 0.6876(-0.0692)     & 0.6324(-0.0970) \\
        No.2 & $\surd$       & $\times$    & $\surd$    & $\surd$      & 0.7032(-0.0536)     & 0.6543(-0.0751) \\
        No.3 & $\surd$       & $\surd$     & $\surd$    & $\times$     & 0.7262(-0.0306)     & 0.6885(-0.0409) \\
        No.4 & $\surd$       & $\surd$     & $\times$   & $\surd$      & 0.7463(-0.0105)     & 0.7043(-0.0251) \\
        No.5 & $\surd$       & $\surd$     & $\surd$    & $\surd$      & \textbf{0.7568}     & \textbf{0.7294} \\
        \bottomrule
    \end{tabular*}
    \footnotesize  
    \label{tab:AS_MC}  
\end{table*}

To evaluate the effectiveness of each component in MEDN and the influence of several key factors on model accuracy, this paper conducts a comprehensive ablation study. Due to its larger data scale and higher proportion of samples with consistent AUs but different emotion categories, the experiments are mainly performed on the CAS(ME)$^3$ dataset. A detailed introduction to the ablation results will be presented subsequently.

\subsubsection{Effect of Proposed Model Components}

Table \ref{tab:AS_MC} presents the ablation results for key components in MEDN, where Experiment No.5 presents the metrics of the complete MEDN method on the CAS(ME)$^3$ dataset. In Experiment No.1, we only adopt a plain ViT as the backbone for the affective feature extraction branch. It can be observed that the UF1 and UAR scores reach 0.6876 and 0.6324 respectively for the three-class classification task on the CAS(ME)$^3$ dataset, indicating a significant performance drop. This demonstrates that our proposed SEVit module achieves superior micro-expression recognition accuracy by enforcing the dispersion of the global attention over static spatial features. In Experiment No.2, we adopt SEVit as the backbone of the affective feature extraction branch, yet only employ a single-scale sparsity rate($s=4$). Although performance is slightly improved compared with Experiment No.1, it still lags considerably behind the optimal results. This indicates that the de-global attention mechanism of SEVit on the spatial dimension is necessary, while multi-scale sparsity rates can further enhance the robustness of the model.

To verify the effectiveness of the proposed CoFM module, in Experiment No.3, we only adopted a simple weighted summation method to fuse the orthogonally decoupled motion and emotion features. The model recognition performance decreased by 0.03-0.04, which demonstrates that in our proposed MEDN method, the collaborative modeling of features by the CoFM module is crucial for improving micro-expression recognition accuracy. In Experiment No.4, we removed the Orthogonal Loss, which also led to varying degrees of performance degradation, especially a 0.0251 drop in UAR. This indicates that orthogonal loss is feasible for the decoupling of explicit motion features and implicit emotion features, providing an important reference for subsequent research on MER.

\subsubsection{Effect of Sparse Rate}

\begin{table}[]
\centering
\setlength{\tabcolsep}{12pt}
\caption{Three-class Classification Results of the MEDN Method on the CAS(ME)$^3$ Dataset with Different Sparsity Rates}
\renewcommand{\arraystretch}{1.2}
\begin{tabular}{cccc}
\hline
Type                            & Sparse Rate          & UF1                  & UAR         \\ \hline
\multirow{4}{*}{Single-scale}   & [1]                  & 0.6873               & 0.6324      \\
                                & [2]                  & 0.7044               & 0.6429      \\
                                & [4]                  & 0.7032               & 0.6543      \\
                                & [8]                  & 0.6986               & 0.6534      \\ \hline
\multirow{3}{*}{Multi-scale}    & [1,2,4]              & 0.7467               & 0.7058      \\
                                & [1,2,4,8]            & \textbf{0.7568}      & 0.7294      \\
                                & [1,1,2,2,4,4,8,8]    & 0.7521               & \textbf{0.7301}      \\ \hline
\end{tabular}
\label{tab:AS_SR}
\end{table}

To evaluate how the sparsity ratio selection of SEVit affects our experiments, we conducted experiments on the CAS(ME)$^3$ dataset using single-scale sparsity ratios and different multi-scale sparsity ratio combinations, respectively. The detailed results are shown in Table \ref{tab:AS_SR}. Experimental results using multi-scale sparsity rates are significantly superior to those using only single-scale sparsity rates, with an improvement of more than 0.04 in the evaluation metric for each comparative experiment. In the single-scale sparsity rate experiments, when $s=1$, the model degenerates into a vanilla ViT and exhibits notably poor performance. The optimal performance is achieved when $s=4,8$. In the multi-scale sparsity rate experiments, the sparsity combination $s=[1,2,4,8]$ yields the best overall performance. Although the configuration $s=[1,1,2,2,4,4,8,8]$ shows no obvious performance gain, it nearly doubles the overall computational cost of the module.

\subsubsection{Impact of different Motion Feature Extraction Backbones}

\begin{table}[]
\centering
\setlength{\tabcolsep}{12pt}
\caption{Results of MEDN on the CAS(ME)$^3$ dataset for the three-class classification task with different AU-Detection backbones.}
\renewcommand{\arraystretch}{1.2}
\begin{tabular}{c|cc}
\hline
Methods                        & UF1                      & UAR             \\ \hline
MEDN + Resnet-18               & 0.7223                   & 0.6789          \\
MEDN + Resnet-34               & 0.7134                   & 0.6625          \\
MEDN + ViT-B/16                & 0.7342                   & 0.6987          \\
MEDN + LED                     & \textbf{0.7568}          & \textbf{0.7294}          \\ \hline
\end{tabular}
\label{tab:AS_MFB}
\end{table}

\begin{figure}
    \centering
        \includegraphics[width=\linewidth]{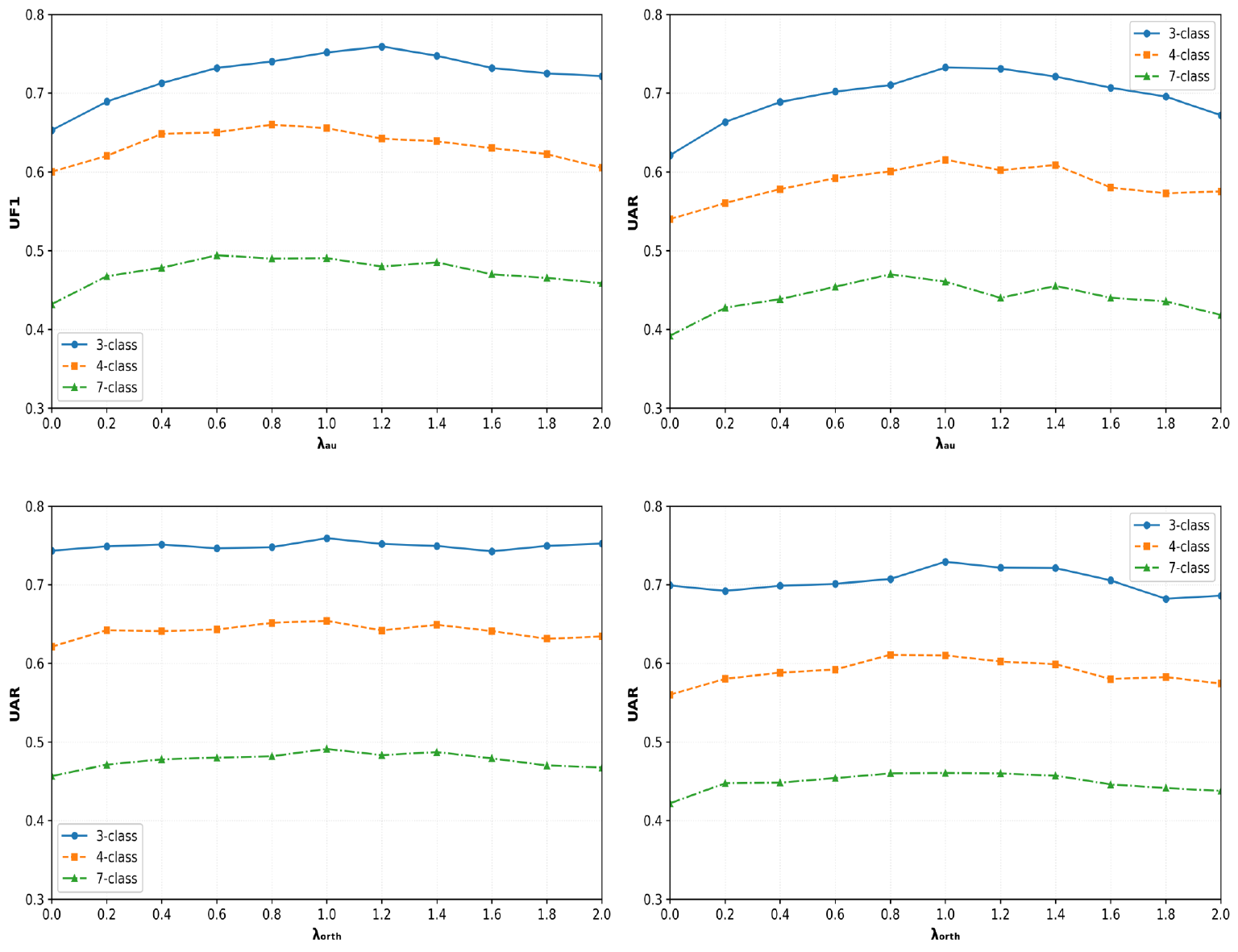}
        \caption{Impact of different $\lambda_{au}$ and $\lambda_{orth}$ on the performance of the MEDN method on the CAS(ME)$^3$ dataset. To better illustrate the performance variations, the vertical axis ranges from 0.3 to 0.8.}
        \label{fig:loss_line}
\end{figure}

As shown in Table \ref{tab:AS_MFB}, to further validate the effectiveness of the MEDN method, we compare the experimental results of mainstream backbone networks with those of the LED model adopted in this work. Specifically, when using ResNet-18/34 and ViT-B/16, we only employ the optical flow between the onset frame and the apex frame as input for the motion detection branch. In contrast, when using LED, we follow the original paper and take the micro-expression video sequence as input directly. It can be observed that using the LED method as the backbone for motion feature extraction yields significant performance advantages. Moreover, this approach ensures that the overall MEDN framework remains independent of apex frame annotations. Particularly, the LED model also exhibits clear advantages in the micro-expression AU-detection task\cite{varanka2023LED}. This indicates that the more accurately explicit motion features are extracted, the more effective the MEDN method becomes, which further verifies the effectiveness of our orthogonal disentanglement of explicit motion features and implicit affective features.

\subsubsection{Selection of $\lambda_{au}$ and $\lambda_{orth}$}

\begin{figure*}[!t]  
    \centering       
    \includegraphics[width=1\textwidth]{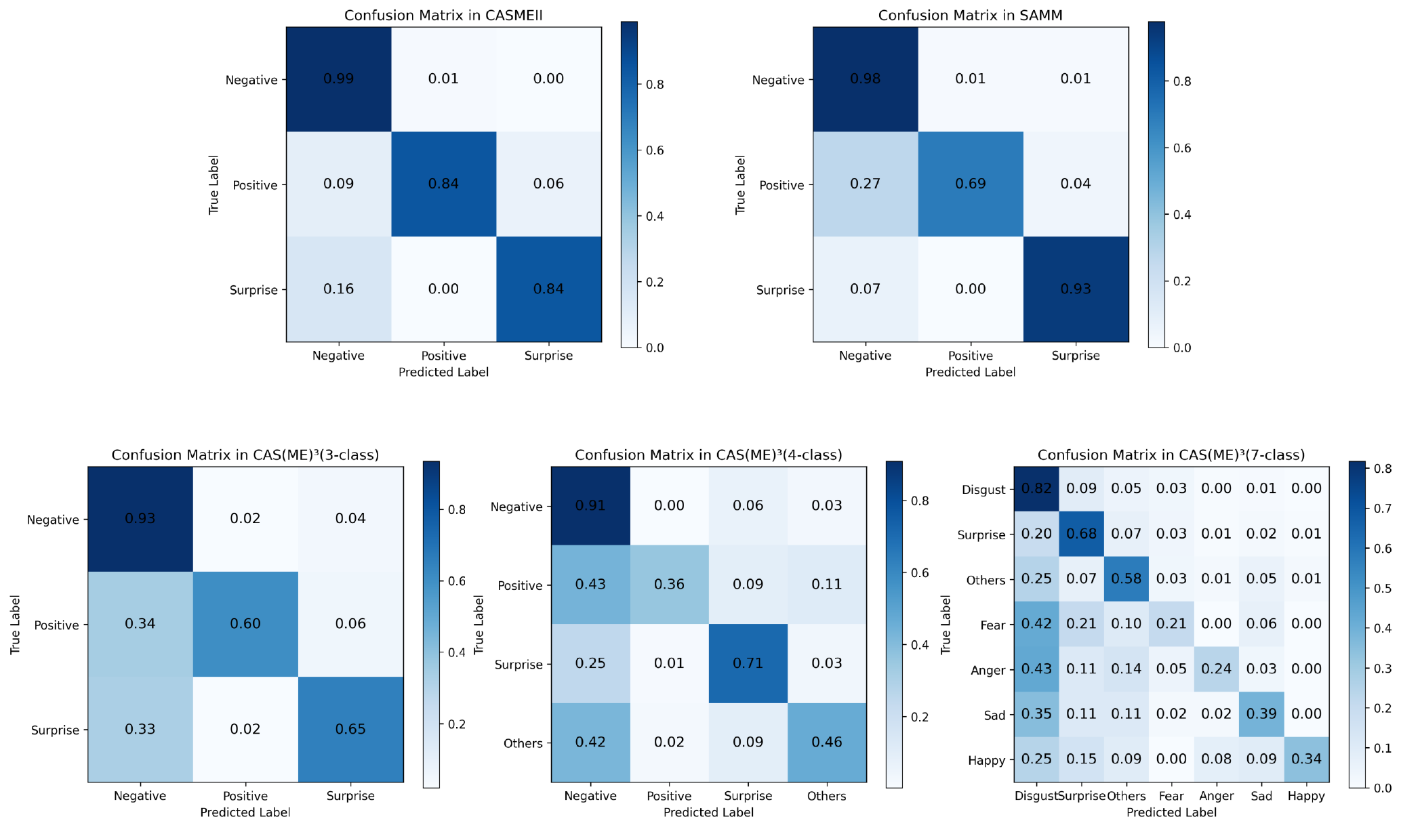}
    \caption{The confusion matrix of our MEDN on the different ME datasets.}
    \label{fig:confusion_matrix}  
\end{figure*}

In our method, MEDN is trained jointly with multiple loss functions, thus introducing two critical hyperparameters, $\lambda_{au}$ and $\lambda_{orth}$. These hyperparameters are used to balance the contribution weights between different losses and exert a significant influence on model training. To determine their optimal values, we conducted a series of comparative experiments, setting the search range for both $\lambda_{au}$ and $\lambda_{orth}$ to \{0.0, 0.2, ..., 2.0\}.

We set $\lambda_{au}$ and $\lambda_{orth}$ to 1.0 respectively, then conduct experiments with different values of the other parameter, and the experimental results are shown in Figure \ref{fig:loss_line}. Experiments show that when $\lambda_{orth}$ is set around 1.0, the model achieves optimal performance across all classification tasks on the CAS(ME)$^3$ dataset. Either excessively low or high values lead to degraded model performance. Compared with $\lambda_{orth}$, $\lambda_{au}$ exhibits higher sensitivity to its value, and the overall optimal performance is obtained within the range of 0.6-1.2. We observe that finer-grained classification tasks tend to correspond to lower optimal $\lambda_{au}$ values, implying a lower reliance on AU-guided explicit motion features. This indicates that in complex micro-expression recognition scenarios, one should not over-rely on explicit motion features, which further validates the overall effectiveness of MEDN.

\subsection{Visual Analysis}

\subsubsection{Confusion matrices}

To further analyze the proposed method, we visualize the confusion matrices of the evaluation results of MEDN on the CASMEII, SAMM, and CAS(ME)$^3$ datasets, as shown in Fig. \ref{fig:confusion_matrix}. It can be observed that MEDN achieves excellent classification performance on all three emotion categories on the CASMEII dataset, especially reaching 0.99 recognition accuracy for the negative category. On the SAMM dataset, MEDN obtains classification accuracy above 0.9 for both the negative and surprise categories, but shows relatively low accuracy for the positive category. This may be attributed to the large individual differences and the extremely small number of positive samples in the SAMM dataset, which makes it difficult for the model to effectively learn the common features of the positive category. On the CAS(ME)$^3$ dataset, MEDN achieves excellent overall accuracy in the 3-class, 4-class, and 7-class classification tasks. However, it can be clearly observed from the figure that the model still tends to focus on emotion categories with more abundant samples. Especially in the 7-class task, the classification accuracy of the dominant category Disgust reaches 0.82, whereas the accuracies for Fear and Anger are only 0.21 and 0.24 respectively. This is because although the CAS(ME)$^3$ dataset provides richer feature representations, the problem of severe class imbalance still exists.

\subsubsection{Feature distribution visualization}

\begin{figure*}[!t]  
    \centering       
    \includegraphics[width=1\textwidth]{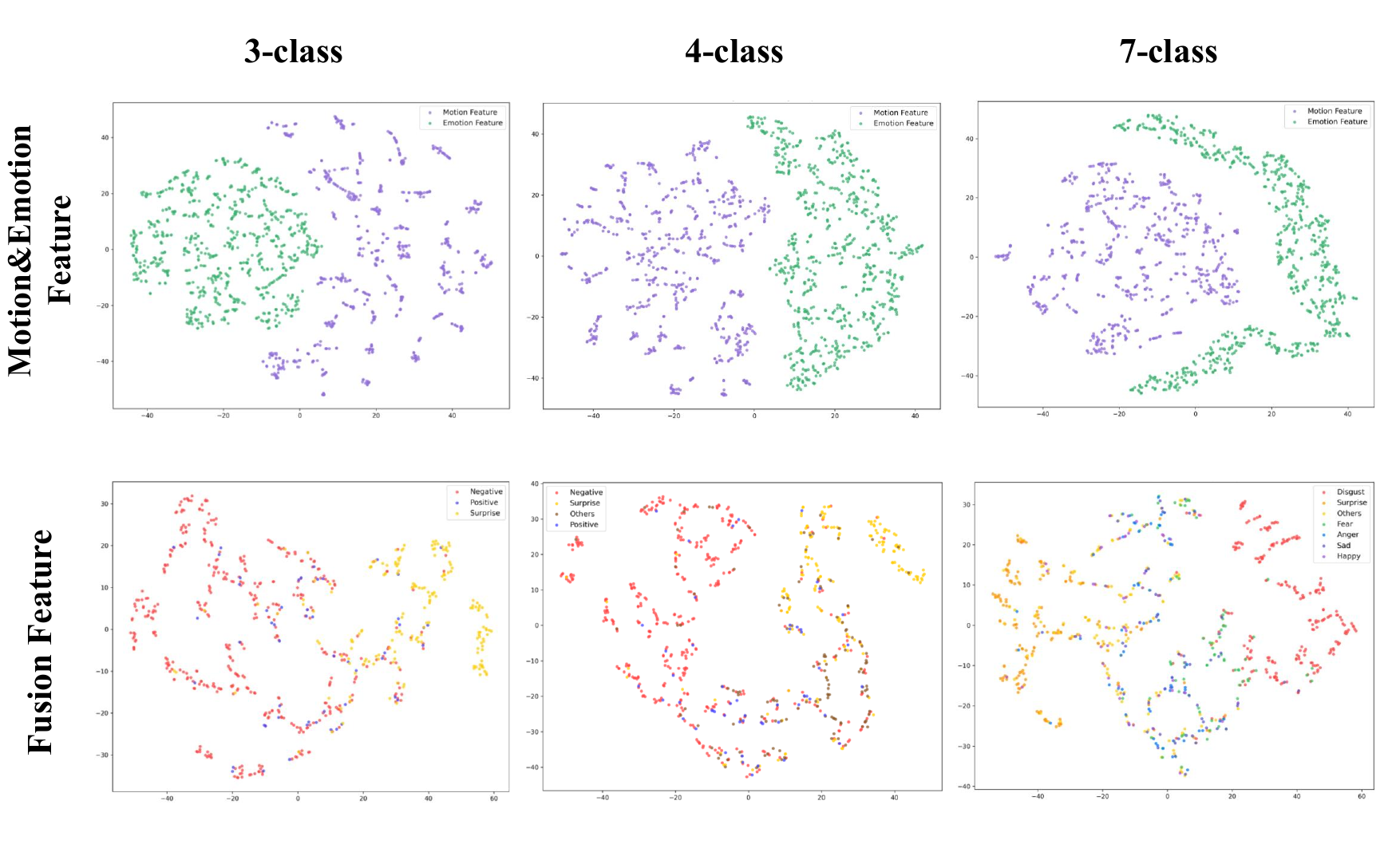}
    \caption{The t-SNE visualization of our MEDN on the CAS(ME)$^3$ datasets.}
    \label{fig:tsne}  
\end{figure*}

For an intuitive demonstration of how the orthogonal loss decouples motion and emotion features, as well as the distribution of fusion features for classification, we employ the t-SNE method to reduce the dimensionality of sample features under different classification tasks on the CAS(ME)$^3$ dataset and visualize them in the form of scatter plots. As shown in Fig. \ref{fig:tsne}, the orthogonal loss enables motion features and emotion features to possess entirely distinct feature spatial distributions, allowing the extraction of emotion features to be independent of explicit motion features. With an increase in category granularity, the explicit motion features guided by AU detection become denser in distribution, while the implicit emotion features exhibit a more discrete and extensive spatial distribution. This indicates that richer emotion categories contain more diverse emotional information, whereas explicit motion features remain globally consistent, further verifying the effectiveness of our method. Although the decoupled motion and emotion features lie in completely separate feature spaces, the figure shows that the aggregated features still maintain relatively clear classification boundaries among categories. This demonstrates that the CoFM module can effectively exploit the collaborative information of decoupled features and fuse the two types of features into a unified feature space, thereby improving MER performance.

\section{Conclusion}
\label{sec:6}
In this paper, we propose a Motion-Emotion Feature Decoupling Network for MER. We explore for the first time the issue of samples with consistent AUs but inconsistent emotional categories in micro-expressions. The motion and emotion features are effectively decoupled via orthogonal loss, and a multi-scale sparse attention mechanism is incorporated to guide the model to focus more on implicit emotional cues beyond explicit motion features. In addition, we perform collaborative fusion on the orthogonally decoupled motion and emotion features. Extensive experiments on multiple datasets demonstrate that MEDN achieves competitive performance compared with state-of-the-art methods, especially in challenging scenarios with high AU-emotion inconsistency.

Despite its promising performance, MEDN still has certain limitations: this method constrains motion feature extraction via facial action unit detection, which relies on the feature extraction capability of the AU detection algorithm itself. Moreover, the training phase requires high-quality action unit annotations, which are costly to label. In the future, we can reduce the dependence on manual annotations through semi-supervised or self-supervised learning, or explore better approaches for explicit motion feature extraction.

\bibliographystyle{IEEEtran}  
\bibliography{mybib}      

\vspace{-52pt}
\begin{IEEEbiography}[{\includegraphics[width=1in,height=1.25in,clip]{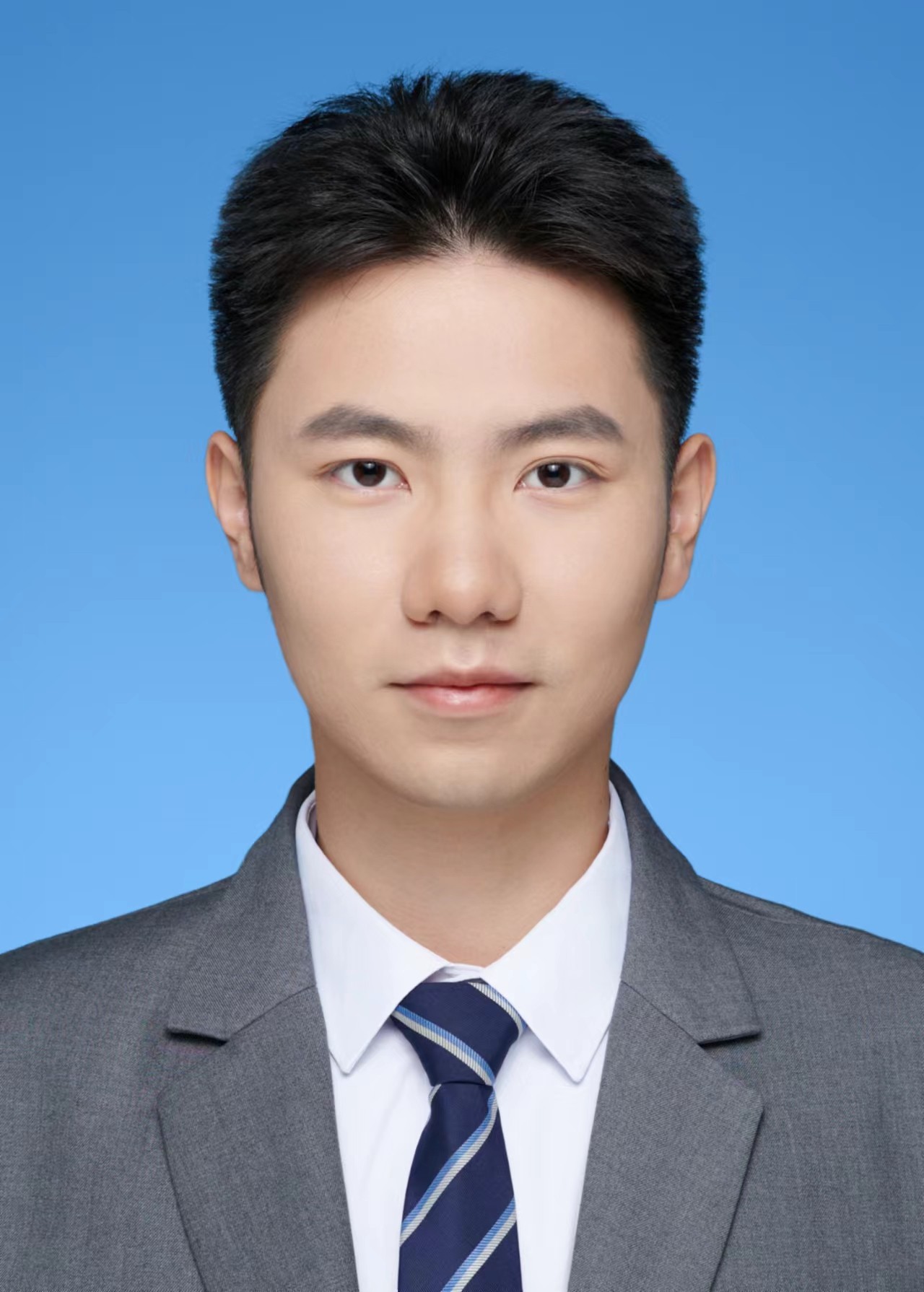}}]{Chenxing Hu}
received his B.S. degree in Computer Science and Technology from Xidian University, Xi'an, China, in 2023. He is currently pursuing his Ph.D. degree at the School of Computer Science and Technology, Xidian University. His research interests include micro-expression analysis, affective computing, and computer vision.
\end{IEEEbiography}

\vspace{11pt}
\begin{IEEEbiography}[{\includegraphics[width=1in,height=1.25in,clip]{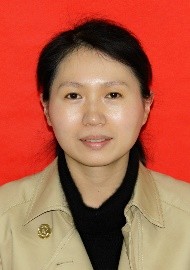}}]{Kun Xie}
received her B.S. degree in Software Engineering and M.S. and Ph.D. degrees in Computer System Architecture from Xidian University in 2003, 2007, and 2014, respectively.  She is currently an professor with the School of Computer Science and Technology, Xidian University.
\end{IEEEbiography}

\vspace{11pt}
\begin{IEEEbiography}[{\includegraphics[width=1in,height=1.25in,clip]{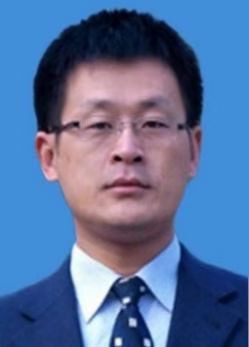}}]{Qiguang Miao}
(Senior Member, IEEE) received the Ph.D. degree in computer application technology from Xidian University, Xi’an, China, in 2005. He is currently a Professor and Ph.D. Student Supervisor with the School of Computer Science and Technology, Xidian University. In recent years, he has authored and coauthored more than 100 papers in significant international journals or conferences. His research interests include intelligent image/video understanding and Big Data.
\end{IEEEbiography}

\vspace{11pt}
\begin{IEEEbiography}[{\includegraphics[width=1in,height=1.25in,clip]{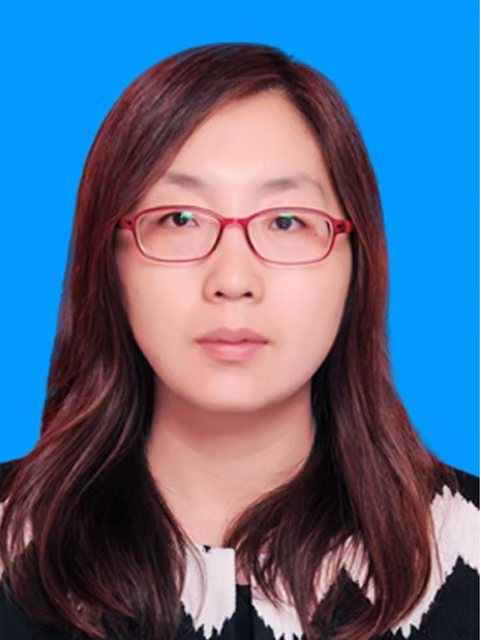}}]{Ruyi Liu}
received the Ph.D. degree in computer application technology from Xidian University, Xi'an, China, in 2018. She is currently working as an associate professor at School of Computer Science and Technology, Xidian University. Her current interests include image classification and segmentation, and computer vision methods with applications in remote sensing.
\end{IEEEbiography}

\vspace{11pt}
\begin{IEEEbiography}[{\includegraphics[width=1in,height=1.25in,clip]{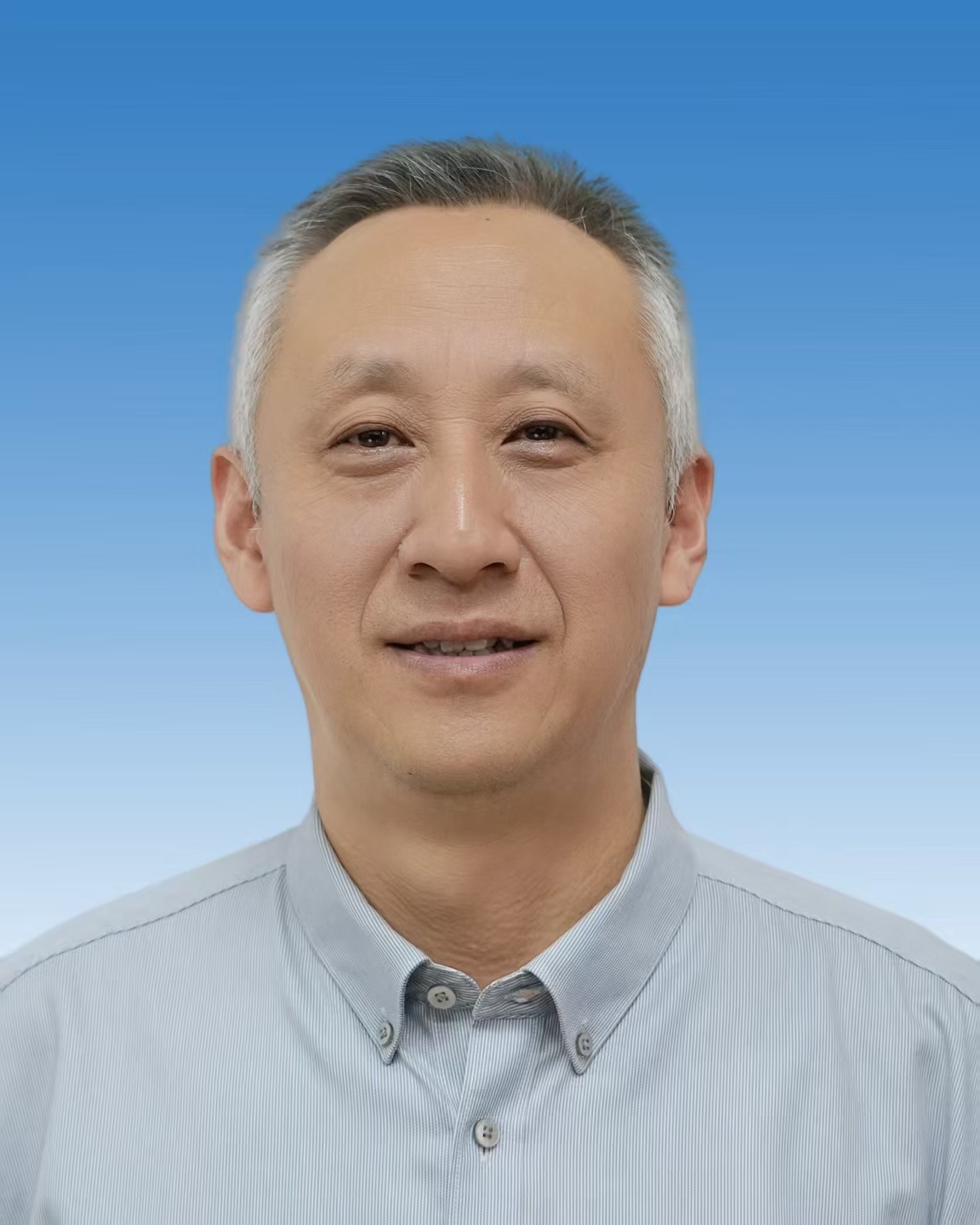}}]{Quan Wang}
received the B.Sc., M.Sc., and Ph.D. degrees in computer science and technology from Xidian University, Xi’an, China, in 1992, 1997, and 2008, respectively, all in computer science and technology. He is currently a Professor with the School of Computer Science and Technology, Xidian University. His research interests include input and output technologies and systems, image processing, and image understanding.
\end{IEEEbiography}

\vspace{11pt}
\begin{IEEEbiography}[{\includegraphics[width=1in,height=1.25in,clip]{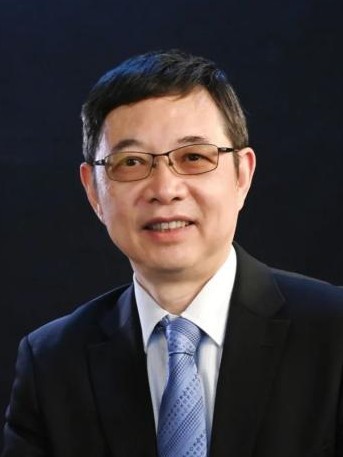}}]{Zongkai Yang}
received the B.S. and M.S. degrees from the Huazhong University of Science and Technology, Wuhan, China, in 1985 and 1988, respectively, and the Ph.D. degree from Xi’an Jiaotong University, Xi’an, China, in 1991. He devoted himself to his postdoctoral research at the Huazhong University of Science and Technology from 1991 to 1993. He is currently a Director and a Professor with the National Engineering Research Center of Educational Big Data, Central China Normal University, Wuhan. His research interests include ICT in education, Management of education.
\end{IEEEbiography}

\vfill

\end{document}